%% file: main.tex
\pdfoutput=1

\documentclass[11pt]{article}

\usepackage[preprint]{acl}

\usepackage{times}
\usepackage{latexsym}

\usepackage[T1]{fontenc}

\usepackage[utf8]{inputenc}

\usepackage{microtype}

\usepackage{inconsolata}

\usepackage{graphicx}
\usepackage{amsmath}
\usepackage{booktabs}
\usepackage{tcolorbox} 
\usepackage{bbm}
 \usepackage{multirow} 
\title{SLMEval: Entropy-Based Calibration for Human-Aligned Evaluation of Large Language Models}

\author{Roland Daynauth, 
  \bf{Christopher Clarke, Krisztian Flautner, 
Lingjia Tang, Jason Mars} \\
University of Michigan \\
\{daynauth, csclarke, manowar, lingjia, profmars\}@umich.edu
  }

\newif\ifsubmission

\newcommand{\mycalink}{
    \ifsubmission
        [Link redacted for review]%
    \else
        https://www.myca.ai%
    \fi
}

\begin{document}
\maketitle
\begin{abstract}

The LLM-as-a-Judge paradigm offers a scalable, reference-free approach for evaluating language models. Although several calibration techniques have been proposed to better align these evaluators with human judgment, prior studies focus primarily on narrow, well-structured benchmarks. As a result, it remains unclear whether such calibrations generalize to real-world, open-ended tasks.

In this work, we show that SOTA calibrated evaluators often fail in these settings, exhibiting weak or even negative correlation with human judgments. To address this, we propose SLMEval, a novel and efficient calibration method based on entropy maximization over a small amount of human preference data. By estimating a latent distribution over model quality and reweighting evaluator scores accordingly, SLMEval achieves strong correlation with human evaluations across two real-world production use cases and the public benchmark. For example, on one such task, SLMEval achieves a Spearman correlation of 0.57 with human judgments, while G-Eval yields a negative correlation. In addition, SLMEval reduces evaluation costs by 5–30× compared to GPT-4-based calibrated evaluators such as G-eval.

\end{abstract}

\input{Introduction}

\input{Related_Work}

\input{Motivation}
\input{Methodology}
\input{Experiments}

\input{Discussion}

\input{Conclusion}

\input{Other}

\bibliography{main}

\input{appendix}

\end{document}

%% file: Introduction.tex
\section{Introduction}

Despite the rapid advancement of large language models (LLMs), reliably evaluating their outputs in a way that aligns with human judgment remains an open challenge. Traditional automated metrics like BLEU~\cite{papineni2002bleu} and ROUGE~\cite{lin2004rouge} offer efficiency and scalability, but dependent on reference outputs and often fail to capture the subjective and contextual nuances that characterize real-world human preferences. Human evaluation remains the gold standard for assessing LLM quality, but is prohibitively expensive, time-consuming, and impractical for large-scale or iterative development cycles \cite{chiang2024chatbot}.

To address these limitations, the LLM-as-a-Judge paradigm has gained widespread adoption as a practical, low-cost alternative to human evaluation~\cite{10.5555/3666122.3668142, liu-etal-2023-g}. However, despite its growing popularity, LLM-based evaluators have certain limitations. Their scores suffer from several well-documented biases and artifacts, including token length bias~\cite{alpaca_eval}, position bias~\cite{wang-etal-2024-large-language-models-fair}, and repeated score patterns~\cite{liu2023g}, all of which can distort comparative judgments and reduce reliability.

To mitigate these issues, a range of calibration techniques have been proposed to better align LLM-as-a-Judge evaluations with human judgments. Yet most prior studies validate these methods on narrow, well-structured benchmarks, such as summarization, translation, and question answering, where task definitions are clear and evaluation criteria are constrained~\cite{yuan2023batcheval, alpaca_eval, liu-etal-2023-g}. As a result, it remains unclear whether these calibration strategies generalize to more subjective, open-ended tasks commonly encountered in real-world deployments.

In this work, we demonstrated that the state-of-the-art calibrated LLM-based evaluators~\cite{yuan2023batcheval, alpaca_eval, liu-etal-2023-g} frequently fail on such open-ended tasks, often yielding weak or even negative correlation with human judgments. To address this limitation, we introduce SLMEval, a novel and effective evaluation framework that uses entropy maximization to calibrate LLM-generated scores based on a small amount of human evaluation data. This calibration promotes well-calibrated LLM-based judgments that better correlate with human evaluations. In addition to better calibration results, unlike prior methods that rely on computationally intensive techniques, such as chain-of-thought prompting or multiple API calls, SLMEval operates using a single-pass evaluation with a small language model (SLM). This dramatically reduces evaluation costs and enables scalable deployment in real-world pipelines, without sacrificing alignment with human evaluations.

We summarize the key contributions of this work as follows: 

\begin{enumerate}
    \item \textbf{Improved Alignment with Human Judgments}: We show that state-of-the-art evaluation methods, such as G-Eval \cite{liu-etal-2023-g} and GPTScore \cite{fu2023gptscore}, perform poorly in subjective real-world tasks. In contrast, our method substantially improves alignment with human judgments. For example, in our production use case, our approach achieves a Spearman correlation of 0.57, while G-Eval yields a negative correlation, highlighting a significant performance gap in this setting.
    \item \textbf{Reduced Computational Overhead}: Using a small language model (SLM), our approach achieves strong performance while significantly reducing the number of required API calls. For example, on a public dataset, our method achieves comparable or better alignment with human judgments while reducing evaluation costs by up to 30× compared to GPT-based evaluators..
    \item \textbf{Evaluation in Real-World and Generalized Settings}: We evaluate SLMEval on two real-world production tasks and a public benchmark to demonstrate its generalizability and practical relevance across diverse evaluation scenarios.
\end{enumerate}


%% file: Related_Work.tex
\section{Related Work}

\paragraph{Traditional Evaluation Methods}
Embedding-based methods such as SBERT \cite{reimers2019sentence} and BERTScore \cite{zhang2020bertscore} improved contextual evaluations, while BLEURT \cite{sellam2020bleurt} fine-tunes models on human-labeled data for better alignment. However, reliance on reference outputs limits their effectiveness in open-ended tasks \cite{liu2024calibrating}.

\paragraph{LLM-as-a-Judge Paradigm}
LLM-based evaluators \cite{liu-etal-2023-g, fu2024gptscore, slam1, kocmi2023large, chen2023exploring}, assess the output without references, reducing the dependency on human annotation. Despite efficiency gains, these models suffer from position and verbosity biases, which impact reliability \cite{dubois2024length}.

\paragraph{Alignment with Human Preferences}
Efforts to align LLM evaluations with human judgments include reinforcement learning with human feedback (RLHF) \cite{ji2024pku, askell2021general} and direct preference optimization (DPO) \cite{rafailov2024direct}. However, these approaches require extensive labeled data and significant computational resources, which limits scalability.

\paragraph{Calibrating Evaluation with Human Preferences}
Studies such as \cite{liu2024calibrating} and \cite{wang-etal-2024-large-language-models-fair} explore human-in-the-loop calibration to improve the performance of LLM-based evaluators. These methods depend on expert judgments, which may not always be feasible, particularly in open-ended tasks.

%% file: Methodology.tex
\section{SLMEval Framework}

\subsection{SLMEval Overview}
Let $M = \{1,\cdots,n\}$ be the set of models under evaluation. For a given prompt $Q$, an LLM or SLM-based evaluator $f$ compares two model outputs $R_i$ and $R_j$ from models $i, j\in M$ where $i\ne j$. The evaluator $f$ returns scalar scores $S_{ij}, S_{ji} \in [1, 10]$ such that:
\begin{equation}
    f(Q, R_i, R_j) = \{S_{ij}, S_{ji}\}
\end{equation}
By default, model $i$ is said to beat $j$ if $S_{ij} > S_{ji}$. However, scores suffer from several well-documented biases and limitations, such as token length bias \cite{alpaca_eval}, position bias \cite{wang-etal-2024-large-language-models-fair}, and repeated score patterns \cite{liu2023g} that can distort comparative judgments. 

To mitigate these biases, we introduce weights $p=\{p_i\}^{n}_{i=1}$ similar to G-Eval \cite{liu2023g} with the constraint
\begin{equation}
\label{eq:sum}
    p_i > 0, \quad \sum_{i=1}^np_i = 1
\end{equation}

Unlike G-Eval, which refines evaluator scores using token-level output probabilities, SLMEval assigns each model $i$  a latent strength parameter $p_i$ estimated from a distribution designed to reflect human preferences. These parameters are then used to calibrate the evaluator's raw scores. The win condition between models $i$ and $j$ is redefined as:

\begin{equation}
    \text{model } i \text{ beats } j \iff p_iS_{ij} > p_jS_{ji}
\end{equation}

\subsection{Entropy-Based Weight Estimation}

To obtain an unbiased estimate of $p$, we maximize the Shannon entropy using the principle of maximum entropy \cite{MR87305}, selecting the distribution with the highest entropy among those consistent with observed human preferences.

\begin{equation}
\label{eq:entropy}
\max_{p} H(p) = -\sum_{i=1}^n p_i \log p_i
\end{equation}

This objective is optimized subject to constraints derived from observed human comparisons. Details of the optimization procedure are provided in Appendix~\ref{sec:max_ent}.






\subsection{Human Preference Constraints}

Let $ \mathcal{D}_{\text{human}} \subseteq M \times M$ denote the set of model pairs for which we have a small amount of sampled human evaluation data for calibration purposes. For each $(i, j)\in \mathcal{D}_{\text{human}}$, we define 

\[
    P(i > j ) = \text{Pr}[\text{human judges prefer } R_i \text{ over } R_j]
\]

If preferences were perfectly consistent and followed a Bradley-Terry model, then $P(i > j) = \frac{p_i}{p_i + p_j}$ \cite{bradley1952rank}. In practice, human data are noisy and may violate transitivity \cite{bradley1952rank, daynauth2024ranking}. To ensure robustness, we use the relaxed constraint:

\begin{equation}
\label{eq:constraint}
    p_i \geq P(i > j)(p_i + p_j), \quad \forall\in \mathcal{D}_{\text{human}}
\end{equation}

\subsection{Optimization and Final Ranking}

SLMEval computes the calibrated weights $p$ by maximizing the entropy function $H(p)$, subject to the normalization constraint (Eq. \ref{eq:sum}) and the relaxed preference constraints (Eq. \ref{eq:constraint}). This yields a probability distribution that reflects the relative strength of each model while remaining minimally biased. 

For each model \( i \in M \), we calculate its win rate by aggregating the results of all calibrated pairwise comparisons against every other model \( j \in M, j \ne i \). Specifically, the win rate is defined as

\[
\text{win\_rate}(i) = \frac{1}{n - 1} \sum_{\substack{j \in M \\ j \ne i}} \mathbbm{1}[p_i S_{ij} > p_j S_{ji}]
\]

The final ranking is obtained by sorting all the models in descending order of \(\text{win\_rate}(i)\).

%% file: Experiments.tex
\section{Experiments}
The application scenario in this study involves the evaluation of a series of generative models (see Appendix \ref{sec:models}) on their response to users of a personal task management and productivity application. Users create and manage their plans and tasks across all aspects of their lives, such as work, personal health, and finances, to stay organized, focused, and productive.

\subsection{Production Use-cases}
Our use-cases focus on generating motivational and actionable advice for users of a to-do list application, which leverages an auto-generative model in the backend to inspire and engage clients. The following two use cases are used:

\begin{itemize}
    \item \textit{Daily Pep Talk (PT)}: At the beginning of each day, an encouraging message is presented to the user based on what they had accomplished the previous day and their goals for the current day.
    \item \textit{Recommendation (RE)} is a personalized suggestion provided to users, helping them organize their tasks into manageable categories. 
\end{itemize}

\subsection{Baselines}
Model responses were ranked using a range of reference-based and LLM-based evaluation metrics, including embedding-based, lexical, and GPT-4-based scorers. A full list of the metrics and corresponding references is provided in Table~\ref{tab:auto_eval_corr_full}. We evaluated two versions of G-Eval—one with probabilistic refinement and one without—both using chain-of-thought reasoning.

\subsection{Implementation Details}
\paragraph{Dataset}

We collect human preference data from 360 Clickworker annotators, each comparing paired model responses for randomly selected prompts from two use cases. The tasks and the model output were uniformly distributed to ensure fairness, yielding ~360 evaluations per task. The prompts were sampled from an internal to-do list application.\footnote{\mycalink}

\paragraph{Evaluator}
For SLMEval, we utilized a 4-bit quantized version of LLaMA 3.1 (with \textit{ temp = 1.0} ) as our evaluation model. This choice was driven by its low computational requirements while retaining much of the performance of its larger counterpart.

%% file: Discussion.tex
\begin{table}[ht]
\centering
\small
\begin{tabular}{lcc}
\toprule
\multirow{2}{*}{Evaluator} & Peptalk & Recommendation \\
                           & $\rho$ & $\rho$ \\
\midrule
USE             & 0.17 & -0.47 \\
TF-IDF          & 0.12 & 0.21  \\
SBERT           & 0.01 & -0.40 \\
BERTScore       & -0.25 & -0.35 \\
BLEURTScore     & 0.20 & -0.47 \\
\midrule
GPTScorer       & 0.35 & -0.15 \\
GPTScore        & -0.14 & 0.39 \\
G-Eval (CoT)    & 0.41 & -0.55 \\
G-Eval (Prob+CoT) & 0.19 & -0.53 \\
GPT-4 + BPC ($k = 3$) & -0.08 & -0.48 \\
\textbf{SLMEval (Ours)} & \textbf{0.48} & \textbf{0.57} \\
\bottomrule
\end{tabular}
\caption{Spearman correlation (\(\rho\)) with human judgments on Peptalk and Recommendation tasks. Bold indicates highest correlation per task.}
\label{tab:auto_eval_corr}
\end{table}

\section{Results and Discussion}

Table~\ref{tab:auto_eval_corr} reports the performance of automated evaluators on each task, measured by the Spearman rank correlation (\(\rho\)). The complete results, including Spearman's \(\rho\) and Kendall's \(\tau\), are provided in Appendix~\ref{sec:full_result}.

\subsection{Performance on Production Use-Case}
GPTScorer aligns reasonably well with human judgments in \textit{Peptalk}, but it does not achieve meaningful alignment in \textit{Recommendation}. In contrast, GPTScore \cite{fu2024gptscore} performs adequately in \textit{Recommendation} but struggles in \textit{Peptalk}. These discrepancies highlight the limitations of current evaluators and emphasize the need for human feedback on tasks that require diverse perspectives.

The negative correlation scores of most evaluators suggest that reference-based evaluation is poorly suited for tasks where multiple diverse responses may be equally valid and human preferences are more subjective. In contrast, SLMEval does not rely on predefined references and outperforms these methods, demonstrating stronger adaptability to real-world tasks with high variability in acceptable output.




\subsection{Performance on Open Datasets}

Although SLMEval is designed for real-world, application-specific tasks, we also evaluate its performance on a standardized open benchmark to assess generalizability. Specifically, we used the FairEval dataset from \cite{wang-etal-2024-large-language-models-fair}, with MT-Bench~\cite{10.5555/3666122.3668142} serving as a human reference. We compare SLMEval’s pairwise prediction accuracy with existing evaluators listed in Table~\ref{tab:auto_eval_corr}.

As shown in Table~\ref{tab:bpc}, SLMEval achieves an accuracy of 58.8\%, outperforming GPTScorer, GPTScore and G-Eval, while approaching GPT-4 + BPC ($k = 3$, 62.5\%)-despite using a smaller quantized model. This result reinforces a broader trend: many evaluators that perform well on structured benchmarks struggle with open-ended tasks, even when drawn from standardized datasets. SLMEval, on the contrary, maintains strong performance in these settings, supporting its robustness beyond application-specific use.

A key advantage of SLMEval is its cost-effectiveness. Although GPT-4 + BPC achieves the highest accuracy, it incurs an API cost approximately six times that of GPTScorer. In contrast, SLMEval uses a 4-bit quantized model served locally via Ollama. When deployed in-house, its operational cost is negligible, as it runs efficiently on a standard laptop. For fair comparison, we estimate its cloud cost using AWS pricing, following the methodology from \cite{slam1}.

\begin{table}[t]
\centering
\small
\scalebox{0.95}{
\begin{tabular}{llc}
\toprule
\textbf{Evaluator}         & \textbf{Accuracy} & \textbf{API Cost Increase}\\
\midrule

GPTScorer  & 52.7\%  &  1.0x\\
GPTScore & 40\%  & 2.0x\\
G-Eval & 45\% & 2.0x\\
GPT4 + BPC (k = 3) & 62.5\% &  6x\\
SLMEval  &  58.8\%   & 0.2x \\

\bottomrule
\end{tabular}}
\caption{Comparison of evaluation accuracy on the FairEval dataset. SLMEval, using a 4-bit quantized model, outperforms GPTScorer and GPTScore while approaching the performance of GPT-4 with BPC ($k=3$).}

\label{tab:bpc}
\end{table}

%% file: Conclusion.tex
\section{Conclusion}

Evaluating the performance of large language models (LLMs) in subjective and application-specific tasks presents unique challenges that traditional automated evaluators often do not address. This paper introduced SLMEval, a scalable and efficient framework designed to bridge the gap between automated evaluation and human preferences.

SLMEval represents a practical and scalable solution to the challenges of evaluating LLMs in subjective tasks, achieving a balance between the reliability of human judgment and the efficiency of automated methods. This improvement is crucial for applications where user satisfaction and perceived utility are paramount, such as providing motivation, advice, or personalized recommendations.

%% file: Other.tex
\section{Limitations}
Our research focuses on a narrow application use case, highlighting the limitations of current autoevaluation techniques. This narrow scope was necessary due to the significant time and monetary investments required to gather human feedback data. Future research will aim to extend this work to more diverse use cases and incorporate traditional benchmarks such as MT-Bench and Chatbot Arena.

Furthermore, future research will explore dynamic recalibration techniques that can adjust evaluation criteria in real time based on ongoing performance metrics and evolving human preferences. These advances have the potential to further enhance the adaptability and responsiveness of SLMEval in a wider range of application scenarios, improving its utility in both subjective and traditional evaluation contexts.

%% file: appendix.tex
\newpage
\appendix

\section{Appendix}

\subsection{Prompts}
\label{sec:prompts}
The prompt consists of two parts: an initial context specific to the use case (see Appendix \ref{sec:peptalk_prompt} for the Peptalk context) and detailed instructions on how scoring should be performed (see Appendix \ref{sec:appendix_prompt} for the evaluation prompt).

\begin{tcolorbox}[colframe=blue!50!black, colback=blue!5!white, coltitle=white, title=SLMEval Prompt Template]
\small
\textit{[User Question]} \\
\textit{$<$prompt$>$} \\

\textit{[The Start of Assistant A’s Answer]} \\
\textit{Model A's Response} \\
\textit{[The End of Assistant A’s Answer]} \\

\textit{[The Start of Assistant B’s Answer]} \\
\textit{Model B's Response} \\
\textit{[The End of Assistant B’s Answer]} \\
\textit{[Evaluator Instructions]} 
\end{tcolorbox}

\subsection{Peptalk Usecase Prompt}
\label{sec:peptalk_prompt}
\begin{tcolorbox}[colframe=black, colback=gray!10, title=Model Prompt]
Imagine you are my personal assistant, generate a short briefing for me at the start of my day. In the briefing, summarize what I completed in the previous day and then give me a preview of the key activities for today. In this briefing, consider my goals for this week and tell me if my focused tasks and rituals are aligned with those goals. Carefully evaluate the associations between the tasks and goals and describe the tasks based on how related you think they are. Note that it is possible that a task is not directly associated with any goals. Reference the specific tasks mentioned in the context and generate this briefing in a single, naturally flowing narrative. Avoid simply listing out tasks one by one. Use a motivating and encouraging tone.
\end{tcolorbox}
\subsection{SLMEval Prompt}
\label{sec:appendix_prompt}
\begin{tcolorbox}[colframe=black, colback=gray!10, title=Evaluator Instructions]
Please act as an impartial judge and evaluate the quality of the responses provided by two AI assistants to the user question displayed below. You should choose the assistant that follows the user’s instructions and answers the user’s question better.

Your evaluation should consider factors such as the clarity, intelligence, likability, trustworthiness, and level of detail of their responses.

Begin your evaluation by comparing the two responses and provide a short explanation. Avoid any position biases and ensure that the order in which the responses were presented does not influence your decision.

Do not allow the length of the responses to influence your evaluation. Do not favor certain names of the assistants. Be as objective as possible.

Each assistant receives an overall score on a scale of 1 to 10, where a higher score indicates a better response.
\end{tcolorbox}


\subsection{LLM/SLM Backend}
\label{sec:models}
The usecases are evaluated using the output of a diverse set of small language models (Table \ref{tab:model_list}) together with GPT-4 \cite{openai_gpt4}.

\begin{table*}[ht]
\label{tab:models}
\centering
\small
\begin{tabular}{lcc}
\toprule
\textbf{Model} & \textbf{Parameter Size} & \textbf{Quantization} \\
\midrule
Zephyr \cite{tunstall2023zephyr} & 7B & 4Bit \\
Mistral \cite{jiang2023mistral} & 7B & 4Bit  \\
StableLM-Zephyr \cite{stablelm-zephyr} & 3B & 4Bit  \\
Starling-LM \cite{starling2023} & 7B &  4Bit  \\
Orca2 \cite{mitra2023orca} & 7B & 4Bit  \\
OpenChat \cite{openllms23} & 7B & 4Bit  \\
LLaMA2 \cite{touvron2023llama} & 7B & 4Bit  \\
Neural-Chat \cite{intel_supervised_finetuning} & 7B & 4Bit  \\
Vicuna \cite{vicuna2023} & 7B & 4Bit  \\
Orca-Mini \cite{mukherjee2023orca} & 3B & 4Bit  \\
\bottomrule
\end{tabular}
\caption{List of evaluated models with their parameter sizes and quantization levels.}
\label{tab:model_list}
\end{table*}

\subsection{Entropy Maximization via the Principle of Maximum Entropy}
\label{sec:max_ent}
We estimate the latent model strength distribution \( p = \{p_1, \dots, p_n\} \) using the \textit{Principle of Maximum Entropy}, which selects the least biased distribution consistent with observed constraints. Specifically, we solve for \( p \) that maximizes the Shannon entropy:
\[
H(p) = -\sum_{i=1}^n p_i \log p_i
\]
subject to:
\begin{enumerate}
    \item \textbf{Normalization:}
    \[
    \sum_{i=1}^n p_i = 1
    \]
    \item \textbf{Empirical preference constraints:}
    \[
    p_i \geq P(i > j)(p_i + p_j), \quad \forall (i, j) \in \mathcal{D}_{\text{human}}
    \]
    where \( P(i > j) \) is the empirical probability that model \( i \) is preferred over model \( j \) based on the human-labeled dataset \( \mathcal{D}_{\text{human}} \).
\end{enumerate}

\subsubsection{Optimization Procedure}
We solve the above as a constrained optimization problem using the \texttt{SLSQP} method in \texttt{scipy.optimize.minimize}. Since \texttt{minimize} performs minimization, we instead minimize the negative entropy:
\[
\min_{p} \quad \sum_{i=1}^n p_i \log p_i
\]
subject to:
\begin{itemize}
    \item \textbf{Equality constraint:} \( \sum_{i=1}^n p_i = 1 \)
    \item \textbf{Inequality constraints:} For each \( (i, j) \in \mathcal{D}_{\text{human}} \),
    \[
    p_i - P(i > j)(p_i + p_j) \geq 0
    \]
    \item \textbf{Bound constraints:} \( p_i \geq \epsilon \) for a small \( \epsilon > 0 \) (e.g., \( 10^{-8} \)) to avoid numerical instability.
\end{itemize}

\subsubsection{Implementation Notes}
\begin{itemize}
    \item The entropy function is strictly concave, and the constraint set is convex, making the problem well-suited to gradient-based solvers.
    \item We initialize \( p \) with the uniform distribution: \( p_i = 1/n \) for all \( i \).
    \item Constraints are enforced with tight numerical tolerances (e.g., \texttt{tol} = \texttt{1e-8}) to maintain solution validity.
\end{itemize}

\subsubsection{Stability Considerations}
To ensure numerical stability:
\begin{itemize}
    \item All \( p_i \) are constrained to be strictly positive.
    \item Logarithmic terms \( \log p_i \) are safe from domain errors due to the positivity constraint.
\end{itemize}

\subsubsection{Full Result}
\label{sec:full_result}
\begin{table*}[ht]
\centering
\small
\begin{tabular}{lcccc}
\toprule
Evaluator & \multicolumn{2}{c}{Peptalk} & \multicolumn{2}{c}{Recommendation} \\
 & $\rho$ & $\tau$ & $\rho$ & $\tau$ \\
\midrule
USE \cite{cer2018universal}          & 0.17 & 0.13 & -0.47 & -0.36 \\
TF-IDF \cite{rajaraman2011mining}    & 0.12 & 0.05 &  0.21 & 0.18  \\
SBERT \cite{reimers2019sentence}     & 0.01 & -0.02 & -0.40 & -0.26 \\
BERTScore \cite{zhang2020bertscore}  & -0.25 & -0.13 & -0.35 & -0.26 \\
BLEURTScore \cite{sellam2020bleurt}  & 0.20 & 0.16 & -0.47 & -0.33 \\
\midrule
GPTScorer \cite{slam1}               & 0.35 & 0.28 & -0.15 & -0.10 \\
GPTScore \cite{fu2024gptscore}                           & -0.14 & -0.07 & 0.39 & 0.29 \\
G-Eval - cot \cite{liu2023g}                           & 0.41 & \textbf{0.35} & -0.55 & -0.4 \\
G-Eval - probs + cot                         & 0.19 & 0.1 & -0.53 & -0.37 \\
GPT4 + BPC ($k = 3$) \cite{wang-etal-2024-large-language-models-fair}                           & -0.08 & -0.10 & -0.48 & -0.56 \\

SLMEval (Ours)                       & \textbf{0.48} & 0.34 & \textbf{0.57} & \textbf{0.42} \\
\bottomrule
\end{tabular}
\caption{Comparison of Correlation Scores ($\rho$: Spearman, $\tau$: Kendall) for Auto Evaluators on Peptalk and Recommendation tasks. Bold values indicate the highest correlation scores for each task.}
\label{tab:auto_eval_corr_full}
\end{table*}